\documentclass[10pt,twocolumn,letterpaper]{article}

\usepackage{cvpr}              
\usepackage{url}
\usepackage{algorithm}
\usepackage{algpseudocode}
\usepackage{makecell}
\usepackage{graphicx}
\usepackage{amsmath}
\usepackage{amssymb}
\usepackage{booktabs}
\usepackage{comment}
\usepackage{threeparttable}
\usepackage{xcolor}
\usepackage{multirow}
\usepackage{verbatimbox}
\usepackage{subcaption}
\newcounter{subtable@save}
\usepackage[export]{adjustbox}
\usepackage[misc]{ifsym}

\definecolor{cvprblue}{rgb}{0.21,0.49,0.74}
\usepackage[pagebackref,breaklinks,colorlinks,citecolor=cvprblue]{hyperref}

\def\paperID{18} 
\def\confName{CVPR}
\def\confYear{2024}

\title{Utilizing Synthetic Data for Medical Vision-Language Pre-training: Bypassing the Need for Real Images}

\author{Che Liu (\Letter)\\
Imperial College London\\
{\tt\small che.liu21@imperial.ac.uk}
\and
Anand Shah\\
Imperial College London\\
{\tt\small s.anand@imperial.ac.uk}
\and
Wenjia Bai\\
Imperial College London\\
{\tt\small w.bai@imperial.ac.uk}
\and
Rossella Arcucci\\
Imperial College London\\
{\tt\small r.arcucci@imperial.ac.uk}
}

\begin{document}
\maketitle
\begin{abstract}
Medical Vision-Language Pre-training (VLP) learns representations jointly from medical images and paired radiology reports. It typically requires large-scale paired image-text datasets to achieve effective pre-training for both the image encoder and text encoder.  The advent of text-guided generative models raises a compelling question: Can VLP be implemented solely with synthetic images generated from genuine radiology reports, thereby mitigating the need for extensively pairing and curating image-text datasets? In this work, we scrutinize this very question by examining the feasibility and effectiveness of employing synthetic images for medical VLP. We replace real medical images with their synthetic equivalents, generated from authentic medical reports. Utilizing three state-of-the-art VLP algorithms, we exclusively train on these synthetic samples. Our empirical evaluation across three subsequent tasks, namely image classification, semantic segmentation and object detection, reveals that the performance achieved through synthetic data is on par with or even exceeds that obtained with real images. As a pioneering contribution to this domain, we introduce a large-scale synthetic medical image dataset, paired with anonymized real radiology reports. This alleviates the need of sharing medical images, which are not easy to curate and share in practice. The code and the dataset can be found in \href{https://github.com/cheliu-computation/MedSyn-RepLearn/tree/main}{https://github.com/cheliu-computation/MedSyn-RepLearn/tree/main}.
\end{abstract}

\section{Introduction}
Significant advancements have been made in the field of medical vision-language pre-training (VLP), particularly in learning visual knowledge from pairs of medical images and reports~\cite{convirt,huang2021gloria,mgca,liu2023m,wan2023med}. These methods have shown remarkable outcomes on a variety of downstream medical vision tasks, leveraging large-scale paired datasets of medical images and reports. However, the acquisition of such large-scale paired image-text datasets involves substantial costs related to data collection and anonymization of both images and text~\cite{khokhar2014quantifying,ficek2021differential,chen2023generative}. This is necessary as all medical data must be de-identified before publishing to comply with privacy and confidentiality regulations~\cite{kaissis2020secure}.

Several research has employed various data augmentation techniques at the image and text levels to increase the number of paired image-text samples~\cite{dong2023maskclip,cascante2023going}.
However, inappropriate data augmentation can lead to the creation of false positive image-text pairs, as these techniques can alter the semantic meaning of medical data~\cite{van2023exploring,li2023frozen}. 
For instance, a medical report might describe `left lung shows opacity', but the corresponding image, after undergoing random rotation/cropping or random grayscale conversion, might not depict the left lung or might display different intensity levels. Furthermore, these methods still necessitate the use of real image-text pair samples, which can limit the generalizability of medical VLP.

Recent advancements in diffusion generative models, such as Stable Diffusion~\cite{ramesh2022hierarchical,rombach2022high,saharia2022photorealistic}, have facilitated the generation of photorealistic images based solely on text or a combination of text and image inputs. RoentGen~\cite{chambon2022roentgen} recently demonstrated that diffusion models can generate medical-style chest X-ray (CXR) images based on medical texts. In this study, we extensively investigate the effects of synthetic images generated by the general domain generative model, SD, and the medical domain-specific model, RoentGen~\cite{chambon2022roentgen}, on medical VLP. Interestingly, methods pre-trained on synthetic images from RoentGen exhibit performance that is comparable to, or even surpasses, those pre-trained on real images for downstream tasks. However, variants pre-trained on synthetic images from SD show a significant decline in performance across all visual tasks.

\section{Methodology}
\subsection{Text-guided Medical Image Generation}
\begin{figure*}[t!]
    \centering
    \includegraphics[width=1.0\textwidth]{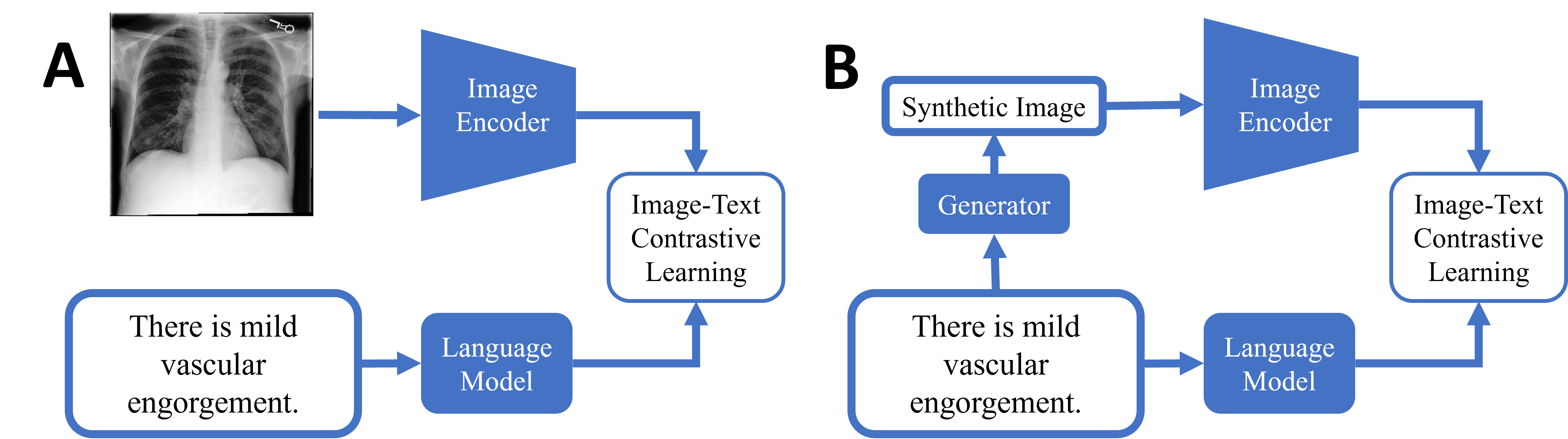}
    \caption{A) Conventional VLP pipeline with real image-text pair. B) Our VLP framework with synthetic image and real medical text. We select SD 2.1 and RoentGen~\cite{chambon2022roentgen} as the generator in this work.}
\label{fig: pipeline}
\end{figure*}

We employ two diffusion-based models in our methodology to generate realistic medical images from actual radiology reports, as depicted in the pipeline illustrated in Fig \ref{fig: pipeline}.

\begin{itemize}
    \item \textbf{Stable Diffusion 2.1 (SD)}~\footnote{https://huggingface.co/stabilityai/stable-diffusion-2-1} is a model proficient in generating images conditioned on text, exhibiting considerable success in intricate tasks of text-guided image generation. However, it is pre-trained on natural image-text datasets and does not specifically target the generation of images in the medical domain.

    \item \textbf{RoentGen}~\cite{chambon2022roentgen} is a model that adapts a pre-trained latent diffusion model to overcome the distributional shift between natural and medical images, generating high-fidelity, diverse synthetic CXR images conditioned on text prompts, including radiology-specific language.
    
\end{itemize}

\begin{figure*}[t!]
    \centering
    \includegraphics[width=1.0\textwidth]{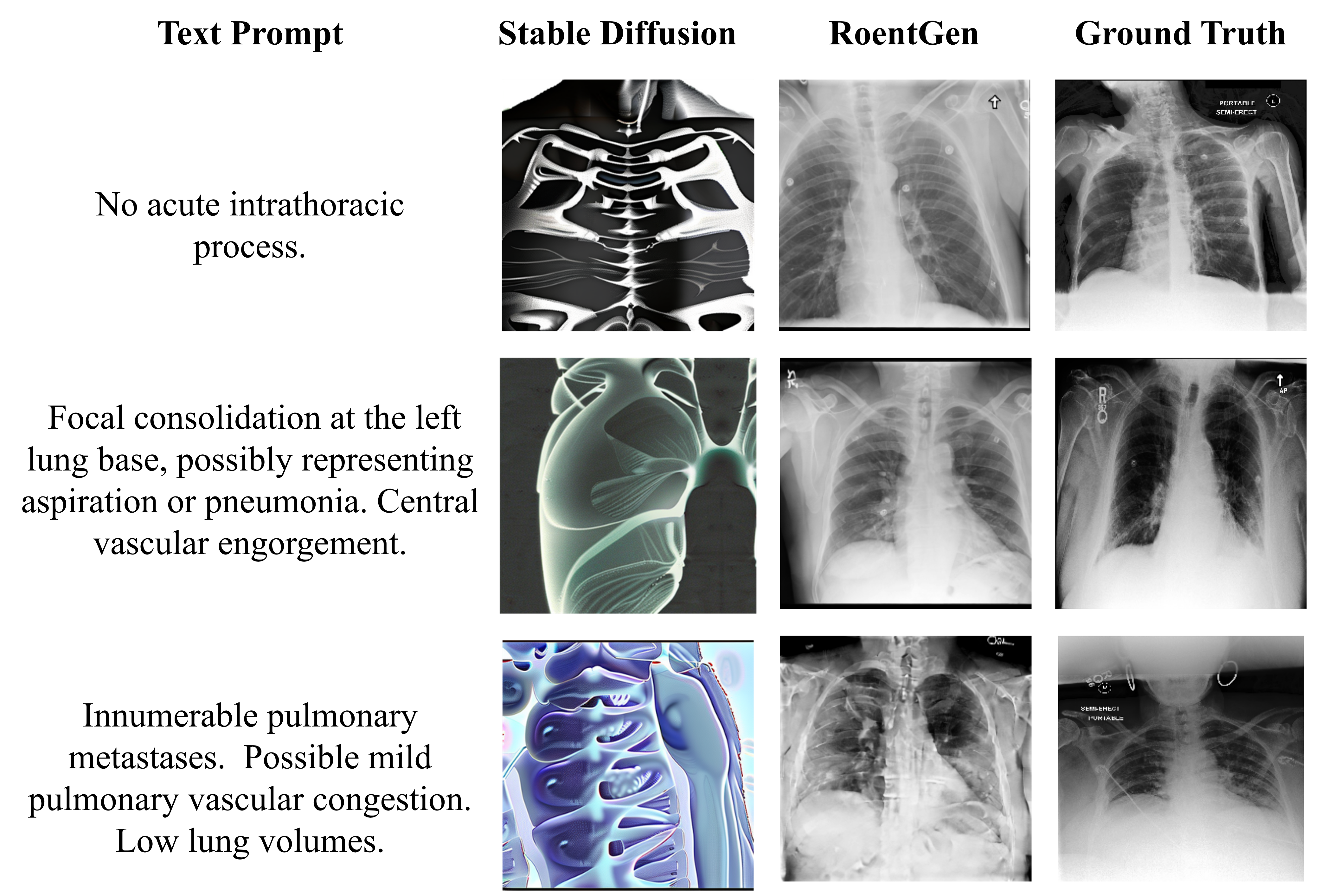}
    \caption{The synthetic samples images generated from real radiology report from SD 2.1 and RoentGen.}
\label{fig: sys}
\end{figure*}

In each stage of image generation, we utilize the `impression' section of the radiology report from the MIMIC-CXR~\cite{johnson2019mimic} dataset as the text condition, with Gaussian noise serving as the initial input for the generative model. We establish the sampling step at 50, the image size at $512\times 512$, and maintain a consistent random seed for all synthetic images. To preserve the one-to-one correspondence between image-text samples, we generate only one synthetic image for each medical report. We select 3 samples from the synthetic dataset produced by both SD and RoentGen for visualization in Fig \ref{fig: sys}, where we also display the corresponding medical prompt and the paired real medical image.

\subsection{Vision-Language Contrastive Pre-training}
The goal of the VLP framework is to derive cross-lingual medical representation from CXR images and their associated radiology reports. We have a training set of $M$ cross-lingual dataset $D \in \mathcal{I} \times \mathcal{T}$, which includes pairs $(i_{m}, t_{m})$. In this case, $\mathcal{I}$ and $\mathcal{T}$ represent the visual and text set, respectively. $i_{m}$ is a raw image and $t_{m}$ is a text report. $m$ is the index of sample and belongs to $M$.

A common VLP architecture primarily consists of an image encoder $\mathcal{E}_{i}: \mathcal{I} \mapsto \mathbb{R}^{D_{i}}$ that encodes the raw image into embeddings of dimension $D_{i}$, and a cross-lingual text encoder $\mathcal{E}_{t}: \mathcal{T} \mapsto \mathbb{R}^{D_{t}}$ that encodes the text report into embeddings of dimension $D_{t}$. Therefore, we have $\mathbf{D}=\left\{\left(\mathbf{i}_{1}, \mathbf{t}_{1}\right),\left(\mathbf{i}_{2}, \mathbf{t}_{2}\right), \ldots, \left(\mathbf{i}_{M}, \mathbf{t}_{M}\right)\right\}$, where $\mathbf{i}_{m} = \mathcal{E}_{i}(i_{m})$ and $\mathbf{t}_{m} = \mathcal{E}_{t}(t_{m})$.

In alignment with the CLIP framework~\cite{clip}, we use a contrastive learning objective to predict the matched pair $(i_{m}, t_{m})$ from $M \times M$ possible image-text pairs, while distancing $M^{2}-M$ negative pairs. Specifically, two non-linear visual and linguistic projectors $\mathcal{P}_{t}$ and $\mathcal{P}_{i}$ are used to transform $\mathbf{i}_{m}$ and $\mathbf{t}_{m}$ into the same dimension $d$, where $\mathbf{\hat{i}}_{m} = \mathcal{P}_{i}(\mathbf{i}_{m})$, $\mathbf{\hat{t}}_{m} = \mathcal{P}_{t}(\mathbf{t}_{m})$, and $\mathbf{\hat{i}}_{m},  \mathbf{\hat{t}}_{m}\in \mathbb{R}^{d}$. 

After obtaining image feature vectors $[\mathbf{\hat{i}}_{m}]_{m=1}^{M}$ and text feature vectors $[\mathbf{\hat{t}}_{m}]_{m=1}^{M}$ from a training batch, we compute cosine similarities $s_{m,m}^{i2t} = \mathbf{\hat{i}}_{m}^{\top} \mathbf{\hat{t}}_{m}$ and $s_{m,m}^{t2i} = \mathbf{\hat{t}}_{m}^{\top} \mathbf{\hat{i}}_{m}$, representing image-text and text-image similarities, respectively. The contrastive vision-language loss $\mathcal{L}_{\mathrm{VLP}}$ is then formulated as follows:

\begin{align}
\vspace{-1.5mm}
\mathcal{L}^{i2t}_{i} = -\log \frac{\mathrm{exp}(s_{m,m}^{i2t} / \sigma_{})}{\sum_{n=1}^{B}{\mathrm{exp}(s_{m,n}^{i2t} / \sigma_{})}},\\ 
\mathcal{L}^{t2i}_{m} = -\log \frac{\mathrm{exp}(s_{m,m}^{t2i} / \sigma_{})}{\sum_{n=1}^{B}{\mathrm{exp}(s_{m,n}^{t2i} / \sigma_{})}}
\end{align}
\begin{equation}
\mathcal{L}_{\mathrm{VLP}}=\frac{1}{2 B} \sum_{m=1}^M\left(\mathcal{L}^{i2t}_{i}+\mathcal{L}^{t2i}_{t}\right), 
\vspace{-0.4mm}
\end{equation}

In the above equations, $\mathcal{L}^{i2t}_{i}$ and $\mathcal{L}^{t2i}_{t}$ are image-text and text-image InforNCE~\cite{Oord2018RepresentationLW} contrastive loss, respectively. $\sigma_{}$ is the temperature hyper-parameter set to 0.07 in our experiments, $B$ is the batch size for each step and $B \in M$. Through the overall loss $\mathcal{L}_{\mathrm{VLP}}$, the model learns maximal mutual information between the matched image-text pairs containing cross-lingual attributes within a batch.

\section{Experiments}
\subsection{Pre-training Details
}
\textbf{Dataset} In this work, we pre-train all baselines on $213,384$ synthetic CXR images with their real medical reports. All medical reports is from MIMIC-CXR dataset.
We pre-process MIMIC-CXR following the approach described in~\cite{convirt,huang2021gloria,mgca}, including image resizing, pixel value normalization, and text tokenization. Additionally, the dataset is filtered by excluding lateral views and reports with less than three tokens. This results in $213,384$ image-text pairs without disease annotation for MIMIC-CXR~\cite{johnson2019mimicjpg}.

\textbf{Implementation}
In our endeavor to investigate the influence of synthetic images on current SOTA methodologies, we have chosen to solely re-implement baseline methods using our synthetic datasets. The implementation and configuration of all pre-training procedures adhere strictly to the official code of the baseline methods, which can be accessed at the following repositories: ConVIRT~\footnote{https://github.com/edreisMD/ConVIRT-pytorch/tree/master}, GLoRIA~\footnote{https://github.com/marshuang80/gloria}, and MGCA~\footnote{https://github.com/HKU-MedAI/MGCA}.
This approach ensures that our exploration remains focused and provides a clear understanding of the impact of synthetic images on these established methods.

\subsection{Downstream Tasks}

\begin{table*}[t!]
\centering
    \vspace{-0.1cm}
    \scalebox{0.65}{
    \begin{tabular}{l|ccc|ccc|ccc}
        \midrule[1.2pt]
         & \multicolumn{3}{c|}{CheXpert (AUC)} & \multicolumn{3}{c|}{RSNA (AUC)} 
        & \multicolumn{3}{c}{COVIDx (ACC)} \\
        Method & 1\% & 10\% & 100\% & 1\% & 10\% & 100\% & 1\% & 10\% & 100\% \\
        \midrule
        Random Init & 56.1 & 62.6 & 65.7 & 58.9 & 69.4 & 74.1 & 50.5 & 60.3 & 70.0 \\
        ImageNet Init & 74.4 & 79.7 & 81.4 & 74.9 & 74.5 & 76.3 & 64.8 & 78.8 & 86.3 \\
        \midrule
        ConVIRT~\cite{convirt} & 85.9 & 86.8 & 87.3 & 77.4 & 80.1 & 81.3 & 72.5 & 82.5 & 92.0  \\ 
        ConVIRT(SD) & 74.6 (\textcolor{red}{$\downarrow$11.3}) & 79.7 (\textcolor{red}{$\downarrow$7.1}) & 82.3 (\textcolor{red}{$\downarrow$5.0}) & 70.1 (\textcolor{red}{$\downarrow$7.3}) & 75.3 (\textcolor{red}{$\downarrow$4.8}) & 77.7 (\textcolor{red}{$\downarrow$3.6}) & 66.5 (\textcolor{red}{$\downarrow$6.0}) & 74.2 (\textcolor{red}{$\downarrow$8.3}) & 84.6 (\textcolor{red}{$\downarrow$7.4}) \\
        ConVIRT(RoentGen) & 84.6 (\textcolor{red}{$\downarrow$1.3}) & 86.4 (\textcolor{red}{$\downarrow$0.4}) & 87.2 (\textcolor{red}{$\downarrow$0.1}) & 78.4 (\textcolor{blue}{$\uparrow$1.0}) & 81.9 (\textcolor{blue}{$\uparrow$1.8}) & 84.7 (\textcolor{blue}{$\uparrow$3.4}) & 70.1 (\textcolor{red}{$\downarrow$2.4}) & 83.6 (\textcolor{blue}{$\uparrow$1.1}) & 92.1 (\textcolor{blue}{$\uparrow$0.1}) \\
        \midrule
        GLoRIA~\cite{huang2021gloria} & 87.1 & 88.7 & 88.0 & 87.0 & 89.4 & 90.2 & 66.5 & 80.5 & 88.8 \\
        GLoRIA(SD) & 78.0 (\textcolor{red}{$\downarrow$9.1}) & 82.7 (\textcolor{red}{$\downarrow$6.0}) & 84.5 (\textcolor{red}{$\downarrow$3.5}) & 79.5 (\textcolor{red}{$\downarrow$7.5}) & 83.8 (\textcolor{red}{$\downarrow$5.6}) & 87.5 (\textcolor{red}{$\downarrow$2.7}) & 62.7 (\textcolor{red}{$\downarrow$3.8}) & 76.5 (\textcolor{red}{$\downarrow$4.0}) & 85.7 (\textcolor{red}{$\downarrow$3.1}) \\
        GLoRIA(RoentGen) & 88.3 (\textcolor{blue}{$\uparrow$1.2}) & 89.6 (\textcolor{blue}{$\uparrow$0.9}) & 89.9 (\textcolor{blue}{$\uparrow$0.9}) & 87.5 (\textcolor{blue}{$\uparrow$0.5}) & 89.8 (\textcolor{blue}{$\uparrow$0.4}) & 90.4 (\textcolor{blue}{$\uparrow$0.2}) & 66.7 (\textcolor{blue}{$\uparrow$0.2}) & 81.0 (\textcolor{blue}{$\uparrow$1.0}) & 89.0 (\textcolor{blue}{$\uparrow$0.2}) \\
        \midrule
        MGCA~\cite{mgca} & 87.6 & 88.0 & 88.2 & 88.6 & 89.1 & 89.9 & 72.0 & 83.5 & 90.5 \\ 
        MGCA(SD) & 80.1 (\textcolor{red}{$\downarrow$7.5}) & 83.5 (\textcolor{red}{$\downarrow$4.5}) & 85.0 (\textcolor{red}{$\downarrow$3.2}) & 82.5 (\textcolor{red}{$\downarrow$6.1}) & 86.5 (\textcolor{red}{$\downarrow$2.6}) & 88.0 (\textcolor{red}{$\downarrow$1.9}) & 68.0 (\textcolor{red}{$\downarrow$4.0}) & 79.5 (\textcolor{red}{$\downarrow$4.0}) & 87.5 (\textcolor{red}{$\downarrow$3.0}) \\
        MGCA(RoentGen) & 88.0 (\textcolor{blue}{$\uparrow$0.4}) & 87.5 (\textcolor{red}{$\downarrow$0.5}) & 87.9 (\textcolor{red}{$\downarrow$0.3}) & 88.7 (\textcolor{blue}{$\uparrow$0.1}) & 89.3 (\textcolor{blue}{$\uparrow$0.2}) & 90.1 (\textcolor{blue}{$\uparrow$0.2}) & 72.5 (\textcolor{blue}{$\uparrow$0.5}) & 83.2 (\textcolor{red}{$\downarrow$0.3}) & 90.3 (\textcolor{red}{$\downarrow$0.2}) \\
        \midrule[1.2pt]
    \end{tabular}
    }
    \caption{Image classification performance on CheXpert, RSNA, and COVIDx datasets with 1\%, 10\%, and 100\% training data. The blue and red colors denote an increase and a decrease in the values respectively compared to the upper bound which uses real images for pre-training.}
    \label{tab:exp_cls}
     \vspace{-0.2cm}
\end{table*}

\begin{table*}[ht!]
    \centering
    \scalebox{0.7}{
    \begin{tabular}{l|c c c| c c c}
     \midrule[1.2pt]
    & \multicolumn{3}{c|}{SIIM (Dice)} & \multicolumn{3}{c}{RSNA (Dice)} \\
    Method & 1\% & 10\% & 100\% & 1\% & 10\% & 100\% \\
    \midrule[1.2pt]
    Random & 9.0 & 28.6 & 54.3 & 6.9 & 10.6 & 18.5 \\
    ImageNet & 10.2 & 35.5 & 63.5 & 34.8 & 39.9 & 64.0 \\
   \midrule[1.2pt]
    ConVIRT & 25.0 & 43.2 & 59.9 & 55.0 & 67.4 & 67.5 \\
    ConVIRT(SD) & 19.6 (\textcolor{red}{$\downarrow$-5.4}) & 38.4 (\textcolor{red}{$\downarrow$-4.8}) & 47.3 (\textcolor{red}{$\downarrow$-12.6}) & 40.1 (\textcolor{red}{$\downarrow$-14.9}) & 52.8 (\textcolor{red}{$\downarrow$-14.6}) & 59.2 (\textcolor{red}{$\downarrow$-8.3}) \\
    ConVIRT(RoentGen) & 26.2 (\textcolor{blue}{$\uparrow$+1.2}) & 43.1 (\textcolor{red}{$\downarrow$-0.1}) & 61.2 (\textcolor{blue}{$\uparrow$+1.3}) & 55.5 (\textcolor{blue}{$\uparrow$+0.5}) & 68.7 (\textcolor{blue}{$\uparrow$+1.3}) & 69.8 (\textcolor{blue}{$\uparrow$+2.3}) \\
    \midrule
    GLoRIA & 37.4 & 57.1 & 64.0 & 60.3 & 68.7 & 68.3 \\
    GLoRIA(SD) & 31.6 (\textcolor{red}{$\downarrow$-5.8}) & 50.9 (\textcolor{red}{$\downarrow$-6.2}) & 57.0 (\textcolor{red}{$\downarrow$-7.0}) & 53.0 (\textcolor{red}{$\downarrow$-7.3}) & 60.8 (\textcolor{red}{$\downarrow$-7.9}) & 61.5 (\textcolor{red}{$\downarrow$-6.8}) \\
    GLoRIA(RoentGen) & 39.1 (\textcolor{blue}{$\uparrow$+1.7}) & 58.4 (\textcolor{blue}{$\uparrow$+1.3}) & 65.5 (\textcolor{blue}{$\uparrow$+1.5}) & 61.5 (\textcolor{blue}{$\uparrow$+1.2}) & 68.5 (\textcolor{red}{$\downarrow$-0.2}) & 69.3 (\textcolor{blue}{$\uparrow$+1.0}) \\
    \midrule
    MGCA & 49.7 & 59.3 & 64.2 & 63.0 & 68.3 & 69.8 \\
    MGCA(SD) & 44.2 (\textcolor{red}{$\downarrow$-5.5}) & 54.0 (\textcolor{red}{$\downarrow$-5.3}) & 60.0 (\textcolor{red}{$\downarrow$-4.2}) & 58.0 (\textcolor{red}{$\downarrow$-5.0}) & 63.0 (\textcolor{red}{$\downarrow$-5.3}) & 64.5 (\textcolor{red}{$\downarrow$-5.3}) \\
    MGCA(RoentGen) & 50.5 (\textcolor{blue}{$\uparrow$+0.8}) & 60.3 (\textcolor{blue}{$\uparrow$+1.0}) & 64.5 (\textcolor{blue}{$\uparrow$+0.3}) & 63.3 (\textcolor{blue}{$\uparrow$+0.3}) & 68.7 (\textcolor{blue}{$\uparrow$+0.4}) & 69.9 (\textcolor{blue}{$\uparrow$+0.1}) \\
    \midrule[1.2pt]
    \end{tabular}
    }
    \caption{Segmentation performance on SIIM and RSNA datasets. The blue and red colors denote an increase and a decrease in the values respectively compared to the no `SD' and no `RoentGen' rows.}
    \label{tab:exp_seg}
\end{table*}

\begin{table*}[ht!]
    \centering
    \scalebox{0.8}{
    \begin{tabular}{l|ccc|ccc}
     \midrule[1.2pt]
    & \multicolumn{3}{c|}{RSNA (mAP)} & \multicolumn{3}{c}{Object CXR (mAP)}\\
    Method & 1\% & 10\% & 100\% & 1\% & 10\% & 100\% \\
    \midrule[1.2pt]
    Random & 1.0 & 4.0 & 8.9 & - & 0.5 & 4.4\\
    ImageNet & 3.6 & 8.0 & 15.7 & - & 2.9 & 8.3 \\
   \midrule[1.2pt]
    ConVIRT & 8.2 & 15.6 & 17.9 & - & 8.6 & 15.9 \\
    ConVIRT(SD) & 4.2 (\textcolor{red}{$\downarrow$-4.0}) & 9.5 (\textcolor{red}{$\downarrow$-6.1}) & 12.4 (\textcolor{red}{$\downarrow$-5.5}) & - & 7.2 (\textcolor{red}{$\downarrow$-1.4}) & 12.8 (\textcolor{red}{$\downarrow$-3.1}) \\
    ConVIRT(RoentGen) & 8.9 (\textcolor{blue}{$\uparrow$+0.7}) & 16.4 (\textcolor{blue}{$\uparrow$+0.8}) & 18.4 (\textcolor{blue}{$\uparrow$+0.5}) & - & 9.2 (\textcolor{blue}{$\uparrow$+0.6}) & 16.4 (\textcolor{blue}{$\uparrow$+0.5}) \\
    \midrule
    GLoRIA & 11.6 & 16.1 & 24.8 & - & 8.9 & 16.6 \\
    GLoRIA(SD) & 5.8 (\textcolor{red}{$\downarrow$-4.8}) & 11.1 (\textcolor{red}{$\downarrow$-5.0}) & 20.2 (\textcolor{red}{$\downarrow$-4.6}) & - & 6.9 (\textcolor{red}{$\downarrow$-2.0}) & 15.3 (\textcolor{red}{$\downarrow$-1.3}) \\
    GLoRIA(RoentGen) & 11.8 (\textcolor{blue}{$\uparrow$+0.2}) & 16.9 (\textcolor{blue}{$\uparrow$+0.8}) & 25.2 (\textcolor{blue}{$\uparrow$+0.4}) & - & 9.2 (\textcolor{blue}{$\uparrow$+0.3}) & 16.8 (\textcolor{blue}{$\uparrow$+0.2}) \\
    \midrule
    MGCA & 12.9 & 16.8 & 24.9 & - & 12.1 & 19.2 \\
    MGCA(SD) & 9.0 (\textcolor{red}{$\downarrow$-5.0}) & 14.0 (\textcolor{red}{$\downarrow$-2.8}) & 22.3 (\textcolor{red}{$\downarrow$-2.6}) & - & 8.0 (\textcolor{red}{$\downarrow$-2.1}) & 16.5 (\textcolor{red}{$\downarrow$-2.7}) \\
    MGCA(RoentGen) & 13.2 (\textcolor{blue}{$\uparrow$+0.3}) & 17.5 (\textcolor{blue}{$\uparrow$+0.7}) & 25.3 (\textcolor{blue}{$\uparrow$+0.4}) & - & 12.3 (\textcolor{blue}{$\uparrow$+0.2}) & 19.6 (\textcolor{blue}{$\uparrow$+0.4}) \\
    \midrule[1.2pt]
    \end{tabular}
    }
    \caption{Object detection performance on RSNA and Object-CXR datasets. The '-' denotes mAP values smaller than 1\%. The blue and red colors denote an increase and a decrease in the values respectively compared to the no `SD' and no `RoentGen' rows.}
    \label{tab:exp_det}
\end{table*}

To thoroughly assess the influence of synthetic images on VLP, we employ three distinct downstream tasks across five diverse CXR image datasets. The image encoder is frozen while the classifier, detector, and decoder are updated for classification, detection, and segmentation tasks, respectively. Except for zero-shot classification, fine-tuning is performed with 1\%, 10\%, and 100\% of the training data, following the data split of MGCA~\cite{mgca}.\\
\textbf{Medical Image Classification}
In the classification task, our objective is to categorize various diseases present in chest X-ray (CXR) images. This task is carried out on three datasets: CheXpert~\cite{irvin2019chexpert}, RSNA~\cite{rsna}, and COVIDx~\cite{wang2020covid}. The CheXpert dataset~\cite{irvin2019chexpert} encompasses five diseases: \textit{atelectasis, cardiomegaly, consolidation, edema,} and \textit{pleural effusion}. The RSNA dataset~\cite{rsna} contains two categories: \textit{normal} and \textit{pneumonia}. The COVIDx dataset~\cite{wang2020covid} comprises three classifications: \textit{COVID-19, non-COVID pneumonia,} and \textit{normal}. For classification, we update only the parameters of a linear layer that has been randomly initialized. The evaluation metrics employed include AUC scores for the CheXpert and RSNA datasets, and accuracy for the COVIDx dataset.\\
\textbf{Medical Image Semantic Segmentation}
In this task, our goal is to segment regions associated with pneumonia and pneumothorax within CXR images. The task employs two datasets, RSNA~\cite{rsna} for pneumonia and SIIM~\cite{siim} for pneumothorax, and utilizes fine-tuned U-Net~\cite{unet} settings for the segmentation process.
The pre-trained vision backbones are frozen encoders, with only the U-Net decoders updated during fine-tuning. Performance is evaluated using Dice scores.\\
\textbf{Medical Image Object Detection}
In this task, our objective is to identify the bounding boxes around abnormal tissues related to pneumonia as well as any foreign objects in CXR images. We accomplish this using the RSNA \cite{rsna} and Object-CXR \cite{obj-cxr} datasets for pneumonia and foreign objects, respectively, and employ YOLOv3 \cite{redmon2018yolov3} as the detection architecture. The pre-trained vision encoder is used as the backbone, with only the detection head updated during fine-tuning. Evaluation is based on Mean Average Precision (mAP) with IOU thresholds 0.4 to 0.75.

\subsection{Baseline Methods}
In order to examine the influence of synthetic images on VLP, we have chosen three SOTA medical VLP methods as our benchmark models:

\begin{itemize}
    \item \textbf{ConVIRT~\cite{convirt}} employs bidirectional contrastive learning to jointly train the vision and text encoders using paired medical images and reports.

    \item \textbf{GLoRIA~\cite{huang2021gloria}} captures the interactions between medical images and reports through both global and regional contrastive learning.

    \item \textbf{MGCA~\cite{mgca}} integrates prototypical contrastive learning~\cite{li2020prototypical} with global and local VLP on paired image-report data.
\end{itemize}

\subsection{Experimental Results}

Tables \ref{tab:exp_cls}, \ref{tab:exp_seg}, \ref{tab:exp_det} delineate the performance of diverse methods on a variety of medical image datasets, spanning three tasks: linear classification, semantic segmentation, and object detection.

Notably, across almost all cases, the use of `RoentGen'~\cite{chambon2022roentgen} consistently improved performance compared to the baseline methods pre-trained on genuine image-text pairs, while a marked decline in performance was observed with the 'SD' approach in comparison to the aforementioned baselines. This observation underscores the potential of synthetic images, generated from actual medical reports via a domain-specific generative model, to enhance the efficacy of an array of medical VLP techniques.

Moreover, a more granular vision-language alignment (e.g., GLoRIA~\cite{huang2021gloria}, MGCA~\cite{mgca}) resulted in improved performance on the synthetic dataset. This suggests that synthetic medical images not only encapsulate global information, but are also endowed with a wealth of localized information that can be specifically aligned with real reports.

\subsection{Impact of Domain-specific Generative Model}
As depicted in Tables \ref{tab:exp_cls}, \ref{tab:exp_seg}, \ref{tab:exp_det}, the baseline methods pre-trained on synthetic medical images from RoentGen~\cite{chambon2022roentgen} exhibit comparable or superior performance relative to those pre-trained on real images. This finding suggests that a domain-specific generative model can yield synthetic image datasets rich in information, enabling VLP models to effectively learn visual representations.

However, it is apparent that the generic generative model, SD 2.1, adversely impacts performance across all tasks and baselines. As shown in Fig \ref{fig: sys} where synthetic samples are displayed, that indicates the difficulty of generic generative models in producing high-fidelity medical images based on medical text prompts, hence restricting their practical applicability in medical contexts.

\section{Conclusion}
In this work, we have conducted the first comprehensive exploration of medical VLP using entirely synthetic medical image datasets paired with real medical reports across multiple vision tasks and five different medical image datasets. Our results demonstrate that domain-specific generative models possess significant potential in generating realistic data for medical VLP. This ability could significantly address the issue of data scarcity in medical VLP and may provide a new way for sharing multi-modal medical datasets, while balancing the risk and benefit in data release. Finally, we are introducing the first large-scale synthetic CXR image dataset that can benefit the research community. Looking ahead, we plan to investigate further applications of text-guided generative models, including data augmentation, zero-shot learning and domain adaptation.

\clearpage
{
    \small
    \bibliographystyle{ieeenat_fullname}
    \bibliography{main}
}


\end{document}